\documentclass[11pt,a4paper]{article}
\usepackage[hyperref]{acl2021}
\usepackage{times}
\usepackage{latexsym}

\usepackage{lipsum}
\usepackage{setspace}
\usepackage{multirow}

\usepackage{amsmath}
\usepackage{amssymb}
\usepackage{mathtools}
\usepackage{booktabs}
\usepackage{tabularx}
\usepackage{multirow}
\usepackage{tikz}
\usepackage{pgfplots}
\usepackage[normalem]{ulem}
\usepackage{xcolor}
\usepackage{dashbox}%
\usepackage{textcomp}
\usepackage{tcolorbox}
\usepackage{adjustbox}
\usepackage{lipsum}
\usepackage[inline]{enumitem}

\pgfplotsset{compat=1.13}

\definecolor{c0}{cmyk}{1,0.3968,0,0.2588} 
\definecolor{c1}{cmyk}{0,0.6175,0.8848,0.1490} 
\definecolor{c2}{cmyk}{0.1127,0.6690,0,0.4431} 
\definecolor{c3}{cmyk}{0.6765,0.2017,0,0.0667} 
\definecolor{c4}{cmyk}{0.3081,0,0.7209,0.3255} 
\definecolor{c5}{cmyk}{0,0.8765,0.7099,0.3647}
\definecolor{darkgrey}{RGB}{180,180,180}
\definecolor{decentgrey}{RGB}{220,220,220}
\usetikzlibrary{calc,fit,positioning,arrows}

\pgfdeclarelayer{bg}
\pgfsetlayers{bg,main}

\tikzset{
	keep name/.style={
		prefix after command={
			\pgfextra{\let\fixname\tikzlastnode}
		}
	},
	partialbox/.style={
		keep name,
		append after command={
			(\fixname.north) -- 
			(\fixname.north west) -- 
			(\fixname.south west) -- 
			([xshift=-#1]\fixname.south)
			(\fixname.north) -- 
			(\fixname.north east) -- 
			(\fixname.south east) -- 
			([xshift=#1]\fixname.south)
		}
	},
	partialbox/.default=5pt
}
\newtcbox{\pattern}{on line,colback=decentgrey!75,colframe=white,size=fbox,arc=3pt, box align=base,before upper=\strut,
top=0pt, bottom=0pt, boxrule=0pt}
\newtcolorbox{multipattern}{on line,colback=decentgrey!75,colframe=white,size=fbox,arc=3pt, box align=base, top=0pt, bottom=2pt, boxrule=0pt, before=\adjustbox{valign=c}\bgroup, after=\egroup, before upper=\strut}

\newcommand\mask{\_\_\_}

\author{
Karen Hambardzumyan$^1$,
~
Hrant Khachatrian$^{1,2}$,
~
Jonathan May$^3$
\\
$^1$YerevaNN,

$^2$Yerevan State University,
\\
$^3$Information Sciences Institute, University of Southern California
\\
  {\tt
  mahnerak@yerevann.com,
  hrant@yerevann.com,
  jonmay@isi.edu
  }
}
\aclfinalcopy 
\def\aclpaperid{4054} 

\title{{WARP}: {W}ord-level {A}dversarial {R}e{P}rogramming}
\begin{document}

\maketitle

\begin{abstract}
Transfer learning from pretrained language models recently became the dominant approach for solving many NLP tasks. A common approach to transfer learning for multiple tasks that maximize parameter sharing trains one or more task-specific layers on top of the language model. In this paper, we present an alternative approach based on adversarial reprogramming, which extends earlier work on automatic prompt generation. Adversarial reprogramming attempts to learn task-specific word embeddings that, when concatenated to the input text, instruct the language model to solve the specified task. Using up to 25K trainable parameters per task, this approach outperforms all existing methods with up to 25M trainable parameters on the public leaderboard of the GLUE benchmark. Our method, initialized with task-specific human-readable prompts, also works in a few-shot setting, outperforming GPT-3 on two SuperGLUE tasks with just 32 training samples. 
\end{abstract}

\section{Introduction}


Language model pretraining has had a tremendous impact on solving many natural language processing tasks \cite{peters-etal-2018-deep, Radford2018ImprovingLU, devlin-etal-2019-bert, Liu2019RoBERTaAR}. 
The most popular two approaches take a pretrained model and use a straightforward supervised learning objective. In the first approach, the parameters of the language model are frozen and a task-specific head is trained on top of them \cite{peters-etal-2018-deep}. The second approach fine-tunes all model parameters \cite{Radford2018ImprovingLU}. The latter can sometimes yield better results \cite{peters-etal-2019-tune}, while the first one usually offers better stability for smaller datasets. The approach based on frozen features does not require storing task-specific language models.

A recent alternative is based on so called adapters \cite{Houlsby2019ParameterEfficientTL, Pfeiffer2020AdapterFusionNT}, a technique that adds new weights at every layer of the pretrained language model while the original parameters are kept frozen. This enables a smaller set of task-specific parameters while achieving results comparable to the fine-tuning approach.

Another approach of leveraging pretrained language models for downstream tasks, introduced by \citet{Radford2019LanguageMA}, provides ``task descriptions'' without using any labeled examples. GPT-3 \cite{Brown2020LanguageMA} demonstrates impressive few-shot learning performance with \textit{priming}: by providing the language model a few inputs and outputs (``analogies'') as a context. The language model contextually ``learns'' from these examples and outputs the answer with a single forward pass without any trainable parameters. These methods, however, require huge language models (1.5B and 175B parameters, respectively).

The success of task reformulation-based approaches suggest that language models are capable of solving various natural language processing tasks given a well-crafted \textit{prompt}. We hypothesize that it is possible to find such prompts. In other words, we can discover extra tokens that, when added to the input, can exploit language model capabilities better than the manually-designed ones.


In this paper, we introduce a novel technique to find optimal prompts.
We call our method \textbf{WARP: Word-level Adversarial RePrograming}\footnote{Our implementation is publicly available at:
\url{https://github.com/YerevaNN/WARP}}. The method is inspired by adversarial reprogramming \cite{Elsayed2018Adversarial} --- a method of adding adversarial perturbations to an input image that reprograms a pretrained neural network to perform classification on a task other than the one it was originally trained for.


We show that our method, using up to 25K trainable parameters per task, achieves $81.6$ test score on the GLUE Leaderboard, outperforming all the other submissions that use up to three orders of magnitude more trainable parameters. 
We show that it is possible to inject knowledge into WARP models using manually designed initialization of the prompt, which is especially useful on tasks with a small number of examples. 
Moreover, WARP shows impressive few-shot performance on two tasks from the SuperGLUE benchmark with just 32 examples, outperforming 
GPT-3 results.
Finally, we discuss the advantages of our method in real-life applications.

\begin{figure}
    \centering
    \includegraphics[width=\columnwidth-85pt]{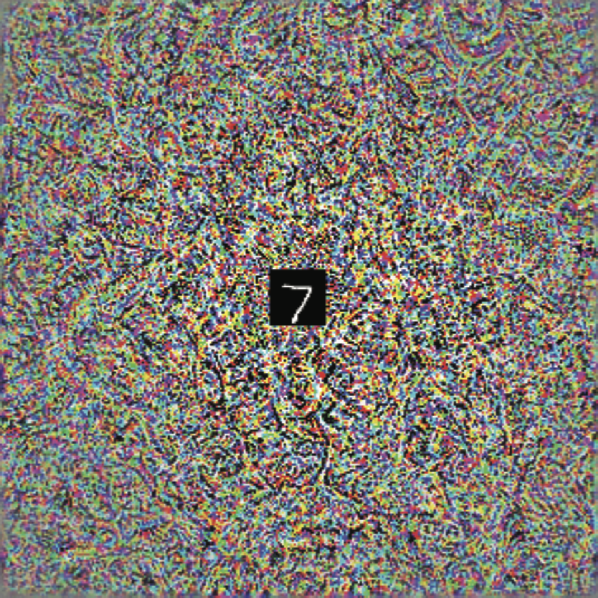}
    \caption{An example of an adversarial program that causes  Inception V3 ImageNet model to function as an MNIST classifier, from \citet{Elsayed2018Adversarial}}
    \label{fig:elsayed_example}
\end{figure}

\section{Related Work}

\subsection{Towards Fewer Trainable Parameters}
\citet{Jiao2020TinyBERTDB} show that knowledge distillation may help reduce the size of their model 7.5 times while almost preserving the performance, but fine-tuning such models still requires storage of separate task-specific models. As seen in Section~\ref{sec:discussion}, this approach does not scale when we want to apply it to many tasks at once.

Another approach, called Adapters \cite{Houlsby2019ParameterEfficientTL, Pfeiffer2020AdapterFusionNT}, introduces new task-specific parameters that are added at every layer of the Transformer network. Only these newly initialized weights are trained, which allows separation of general and task-specific knowledge. In contrast, our method does not inject task-specific knowledge inside the body of the pretrained language model. Instead, it focuses on learning task-specific input-level prompts.


\subsection{Task Reformulation}

\begin{figure}
    \centering
    \includegraphics[width=\columnwidth]{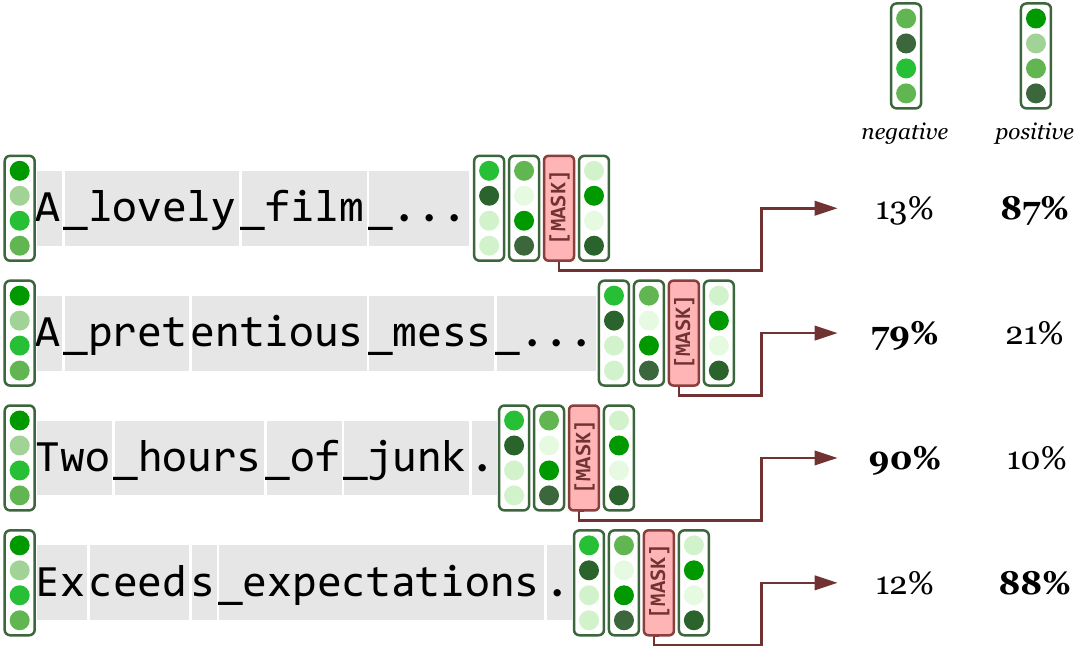}
    \caption{WARP adds a few trainable embeddings around the input, which causes the masked language model to predict the sentiment of the sentence.}
    \label{fig:advnlp}
\end{figure}

In GPT-2, \citet{Radford2019LanguageMA} introduce a completely unsupervised way for transferring knowledge to downstream tasks by reformulating various natural language understanding tasks into language modeling problems.
This approach does not make use of the available training examples. 
\citet{Brown2020LanguageMA} demonstrate an effective few-shot transfer by reformulating downstream tasks into input-output analogies in the context without a need for further fine-tuning. Nonetheless, the number of training examples is limited to the context size and is not scalable to a traditional supervised learning scenario. 

\citet{Schick2020ItsNJ} show the effectiveness of reformulating a number of tasks into Cloze-style tasks by fine-tuning masked language models \cite{devlin-etal-2019-bert}. The method, called Pattern Exploited Training (PET), additionally uses training samples and performs few-shot learning even without huge models such as GPT-3.

Our method is also based on masked language models, but unlike PET, we focus on finding the best prompt using the training examples. This eliminates the need for manually-designed prompts, however, our method can also benefit from similar prior knowledge about the task by careful initialization of the prompts.

\begin{figure*}
 \centering
    \includegraphics[width=400pt]{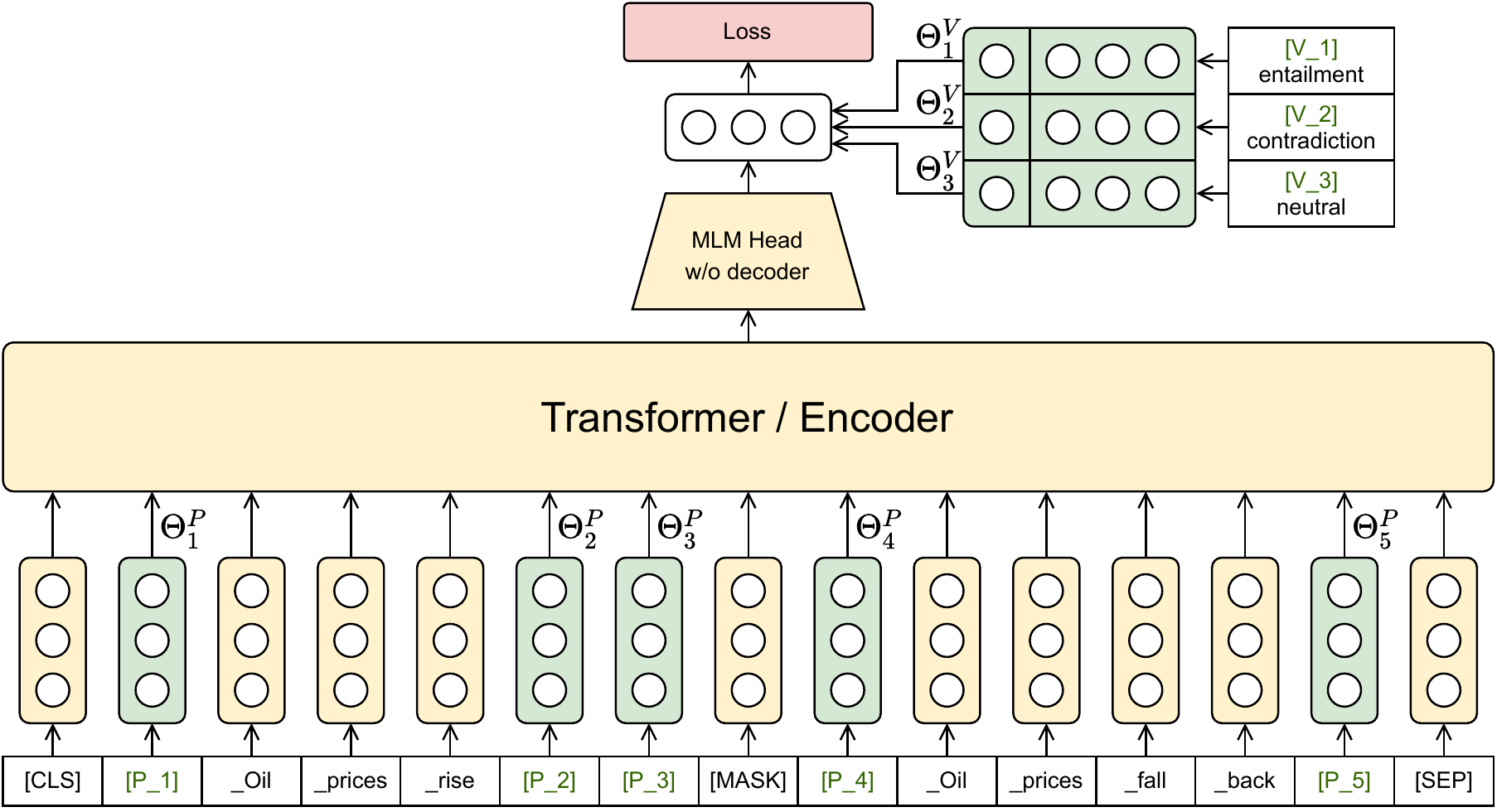}
 \caption{Illustration of WARP. The prompt tokens \texttt{[P\_1]}, \texttt{[P\_2]}, ..., \texttt{[P\_N]} are inserted before, between, and after the sentences. Only the prompt and class embeddings are trainable (colored in green). The masked language modeling Head is applied without the decoder; instead, the matrix of \texttt{[V\_1]}, \texttt{[V\_2]}, ..., \texttt{[V\_N]} is applied as a linear layer. Finally, a regular task-specific loss is computed on the resulting logits.}
 \label{fig:figure-label}
\end{figure*}

\subsection{Adversarial Reprogramming}
Adversarial Reprogramming \cite{Elsayed2018Adversarial} demonstrates the reprogramming of pretrained ImageNet classifiers by adding input-level adversarial perturbations to make them perform well on MNIST and CIFAR-10 image classification tasks. The adversarial perturbation is designed to be image padding added to the original input, as illustrated in Figure~\ref{fig:elsayed_example}. Then the perturbation parameter is trained to optimize the target classification task objective using the annotated image data.

While in the case of image classification it is not obvious why adversarial reprogramming should ever work, e.g. why a network trained on ImageNet should have the capacity to solve MNIST when surrounded with a particular bitmap, for NLP tasks, there is more intuition. Many NLP tasks can be reformulated as language models, a shared space for both program and data.

Adversarial reprogramming has been adapted to text classification tasks with LSTM networks in \cite{adv-reprog-text-lstm}. They operate in the vocabulary space and reprogram a model trained for one task to perform another task. More recently, AutoPrompt \cite{Shin2020AutoPromptEK} attempts to find prompts for large language models automatically without adding any parameters to the model. Unlike AutoPrompt, we perform gradient-based optimization in the space of word embeddings which gives our model more degrees of freedom and eventually better performance on the downstream tasks (Section~\ref{sec:autoprompt}).

In a more general sense, guiding an NLP model with special tokens appended to the input is an even older idea. In particular, multilingual neural machine translation models use special tokens in the input to control the target language \cite{towards-universal-enc-dec,google-multilingual-nmt} or politeness of the translation \cite{controlling-politeness-nmt}. Another method to reprogram a BERT-based model is proposed by \citet{x-lingual-transfer-repr}, where a model tuned on an English version of a particular task is transformed to work in another language by changing only the embedding matrices. 

In parallel work, \citet{prefix-tuning} propose a similar method and successfully apply it on two text generation tasks. Apart from the different types of tasks and our characterization of the task as a form of Adversarial Reprogramming, the main difference between their approach and ours is that they use an additional parameterization trick to stabilize the training.

\section{WARP}
We follow a setup similar to \citet{Elsayed2018Adversarial} with some NLP-specific modifications depicted in Figure \ref{fig:advnlp}. 

Our goal is to find the best prompt that will make a pretrained masked language model predict the desired answer (verbalizer token) for a training example's  masked token\footnote{This approach can be easily extended to autoregressive language modeling.}.
We search for such prompts in the (continuous) embedding space. In other words, we want to find parameters $\Theta = \{\Theta^P, \Theta^V\}$ for prompt and verbalizer embeddings, respectively,  such that:
$$\Theta^* = \arg\max\limits_\Theta\left({-\log{P_\Theta(y|x)}}\right)$$
and the probabilities are given by:
$$P_\Theta(y|x) = \dfrac{
    \exp{\Theta^V_y}  f(T_{\Theta^P}(x))
}{
    \sum\limits_{i \in C}{\exp{\Theta^V_i} f(T_{\Theta^P}(x))}
}$$
where $T_{\Theta^P}(x)$ is the template that inserts the prompt embeddings $\Theta^P$ into predefined positions, $C$ is the set of classes, and $f(x)$ is the masked language model output (without the last decoder layer, which is simply the transposed word embedding matrix). Both $\Theta^P$ and $\Theta^V$ are vectors in the same embeddings space as the word embeddings.

In Figure \ref{fig:advnlp}, the template $T_{\Theta^P}(x)$ prepends $\Theta^P_1$ and appends $\Theta^P_2$, $\Theta^P_3$, $\Theta^P_4$ parameters to the word embeddings and uses $\Theta^V_+$ and $\Theta^V_-$ to calculate the probabilities on the masked token position for positive and negative classes.

\subsection{Method}
Similar to \citet{Elsayed2018Adversarial}, we employ stochastic gradient descent to find the best adversarial perturbation on the text that will minimize the task objective. First, we insert special prompt tokens \texttt{[P\_1]}, \texttt{[P\_2]}, ... \texttt{[P\_K]} and an additional \texttt{[MASK]} token into the input sequence. These tokens might be placed before or after the sentences,  depending on the prompt template.

We set the optimization objective to a cross-entropy loss between the head output of the masked language model and the verbalizer tokens \texttt{[V\_1]}, \texttt{[V\_2]}, ...,  \texttt{[V\_C]} for classes $1...C$ accordingly.

The only trainable parameters are the word embeddings for \texttt{[P\_1]}, ..., \texttt{[P\_K]} and \texttt{[V\_1]}, ... \texttt{[V\_C]}. In case we want to train models for multiple tasks, these are the only task-specific parameters we need to store. The entire ``body'' of the large language model (all attention layers, feed-forward layers, and all other word embeddings) remains untouched. 

Note that, unlike most adversarial attacks, we do not update the embeddings of the original tokens of the input. This follows the intuition from \citet{Elsayed2018Adversarial}, when the pixels of MNIST or CIFAR images are left untouched, and only padding pixels are updated. 

We train these parameters by minimizing the loss on the training set of the downstream task.

\subsection{Implementation Details}

WARP is implemented in the AllenNLP framework. 
For all the GLUE benchmark tasks we use the \texttt{roberta-large} \cite{Liu2019RoBERTaAR} model from the \texttt{PyTorch} implementation of \texttt{huggingface transformers} \cite{wolf-etal-2020-transformers} library. For the few-shot experiments, we use \texttt{albert-xxlarge-v2} in order to directly compare to iPET \cite{Schick2020ItsNJ}. For the GLUE and SuperGLUE tasks we use dataset loaders and metrics implementations from the \texttt{huggingface datasets} library.

The prompt tokens are initialized either with word embeddings of \texttt{[MASK]} or similar to the vectors from the word embedding layer. 
For the answer prompts, we use the masked language model head, which usually consists of a feed-forward network and a decoder on top of it, where the weights of the decoder are shared with the word embeddings used for the input. We calculate the softmax over the verbalizer tokens \texttt{[V\_1]}, ... \texttt{[V\_C]}.

We choose the Adam optimizer with a slanted triangular schedule for the learning rate with 6\% warm-up steps and train for $10$-$20$ epochs on each task.
Each batch consists of examples containing at most $1024$ tokens and $8$ examples.

In order to speed up the training, we disable the dropout of the pretrained language model.
All the experiments are performed on two Titan Vs and two RTX 3080 GPUs, with mixed precision training. In practice, WARP is 2.5-3 times faster than regular fine-tuning and 2 times slower than frozen-features experiments in terms of epoch duration with the  same batch sizes.

Details about the hyperparameters can be found in the Supplementary material. 

\begin{table*}

\centering
\small
\onehalfspacing
\setlength{\tabcolsep}{4.6pt}
\begin{tabular*}{\textwidth}{@{}lcccccccc|cr}
\hline

& \textbf{MNLI}
& \textbf{QNLI}
& \textbf{QQP}
& \textbf{RTE}
& \textbf{SST}
& \textbf{MRPC}
& \textbf{CoLA}
& \textbf{STS-B}
& \textbf{AVG}
& {\textbf{\#} }
\\
\hline


Human Baselines
& $92.0$ / $92.8$ & $91.2$ & $59.5$ / $80.4$ & $93.6$ & $97.8$ & $86.3$ / $80.8$ & $66.4$ & $92.7$ / $92.6$
& $87.1$
&\\
\hline

DeBERT\footnote{Current State of the Art} 
& $91.9$ / $91.6$ & $99.2$ & $76.2$ / $90.8$ & $93.2$ & $97.5$ & $94.0$ / $92.0$ & $71.5$ & $92.9$ / $92.6$
& $90.8$
&$3\cdot 10^{9}$ \\
RoBERTa
& $90.8$ / $90.2$ & $95.4$ & $74.3$ / $90.2$ & $88.2$ & $96.7$ & $92.3$ / $89.8$ & $67.8$ & $92.2$ / $91.9$
& $88.1$
&$355 \cdot 10^{6}$ \\

\hline

BERT\textsubscript{large}
& $86.7$ / $85.9$ & $92.7$ & $72.1$ / $89.3$ & $70.1$ & $94.9$ & $89.3$ / $85.4$ & $60.5$ & $87.6$ / $86.5$
& $80.5$
&$355 \cdot 10^{6}$ \\
BERT\textsubscript{base}
& $84.6$ / $83.4$ & $90.5$ & $71.2$ / $89.2$ & $66.4$ & $93.5$ & $88.9$ / $84.8$ & $52.1$ & $87.1$ / $85.8$
& $78.3$
&$110 \cdot 10^{6}$ \\
TinyBERT\textsubscript{6}
& $84.6$ / $83.2$ & $90.4$ & $71.6$ / $89.1$ & $70.0$ & $93.1$ & $87.3$ / $82.6$ & $51.1$ & $85.0$ / $83.7$
& $78.1$
&$67 \cdot 10^{6}$ \\
TinyBERT\textsubscript{4}
& $82.5$ / $81.8$ & $87.7$ & $71.3$ / $89.2$ & $66.6$ & $92.6$ & $86.4$ / $81.2$ & $44.1$ & $81.9$ / $80.4$
& $75.9$
&$15 \cdot 10^{6}$ \\
ELECTRA\textsubscript{small}
& $81.6$ / $81.2$ & $88.3$ & $70.4$ / $88.0$ & $63.6$ & $91.1$ & $89.0$ / $84.9$ & $55.6$ & $85.6$ / $84.6$
& $77.4$
&$14 \cdot 10^{6}$ \\
Adapters (BERT)
& $85.4$ / $85.0$ & $92.4$ & $71.5$ / $89.4$ & $71.6$ & $94.3$ & $88.7$ / $84.3$ & $59.2$ & $87.3$ / $86.1$
& $80.2$
&$1.2 \cdot 10^{6}$ \\

\hline

WARP (RoBERTa)
& $88.0$ / $88.2$ & $93.5$ & $68.6$ / $87.7$ & $84.3$ & $96.3$ & $88.2$ / $83.9$ & $53.9$ & $89.5$ / $88.8$
& $81.6$
&$<25K$ \\

\hline

\end{tabular*}

\caption{Test set results on GLUE Benchmark. The results are obtained from the GLUE Evaluation server.  The subscript next to TinyBERT corresponds to the number of layers in the model. WARP for RTE, STS-B and MRPC are intialized from the MNLI parameters. Results for WNLI  are not shown, although they are counted in the averaged GLUE score (AVG column).
The last column \textbf{\#} shows the number of \textit{trainable} parameters. WARP's average performance is higher than all models with up to three orders of magnitude more trainable parameters. Fully fine-tuned RoBERTa and the current state-of-the-art method (DeBERT) score higher by 6.5 and 9.2 points, respectively. 
}
\label{tab:gluetest}

\end{table*}

\section{Experiments on GLUE}
\label{section_glue}

Following prior work, we evaluate our method on the GLUE Benchmark \cite{wang2019glue}, which consists of 9 natural language understanding tasks. Generally, we perform single-task WARP training, with early stopping and model selection using the original validation sets, if not stated otherwise.

\subsection{Tasks}
\label{section_glue_tasks}

Almost all the tasks from the GLUE Benchmark are either sentence classification or sentence pair classification tasks, so WARP requires very few modifications to adapt to each of the tasks.

\textbf{SST-2} \citep[Sentence Sentiment Treebank,][]{tasks_glue_sst} is a single sentence binary classification task. For the prompt, we put a \texttt{[MASK]} token after the sentence, and the trainable prompt tokens are both appended and prepended to the sentence.

\textbf{CoLA} \citep[Corpus of Linguistic Acceptability,][]{tasks_glue_cola} is a single sentence classification task as well, so we treat both the same way with the only difference that as a validation metric we use accuracy for SST-2, and Matthew's correlation for CoLA.

\textbf{MNLI} \citep[MultiNLI, Multi-Genre Natural Language Inference,][]{tasks_glue_mnli}, \textbf{QNLI} \citep[Question Natural Language Inference,][]{tasks_glue_qnli} and \textbf{RTE} \citep[Recognizing Textual Entailment,][]{tasks_glue_rte_1,tasks_glue_rte_2,tasks_glue_rte_3,tasks_glue_rte_4} are sentence pair classification tasks. Similar to \citet{Schick2020ExploitingCQ}, we may have prompt tokens before, after and between the two sentences, but the \texttt{[MASK]} token is always put between the sentences. 
For MNLI, we use matched accuracy as a validation metric and use the same model for the mismatched version. In our few-shot attempt for the RTE task, we use a different training and evaluation setup discussed in Section~\ref{section_few_shot_tasks}.
\textbf{QQP} (Quora Question Pairs\footnote{https://www.quora.com/q/quoradata/First-Quora-Dataset-Release-Question-Pairs})  and \textbf{MRPC} \citep[Microsoft Research Paraphrase Corpus,][]{tasks_glue_mrpc} follow the same prompt pattern as NLI tasks. As a validation metric $F_1$ score is used.

\textbf{STS-B} \citep[Semantic Textual Similarity Benchmark,][]{tasks_glue_stsb}, unlike the other tasks in the benchmark, is formulated as a regression task. The prompt pattern is the same, but instead of introducing new embeddings for \texttt{[V\_1], [V\_2], ..., [V\_C]} verbalizer tokens, we add a regression head to the last hidden state of MLM head and use Mean Squares Error optimization objective, similar to \cite{Liu2019RoBERTaAR}. Pearson Correlation is used as the validation metric. During inference, we clip the scores within $[1, 5]$.

We follow \citeauthor{Liu2019RoBERTaAR} and train models for MRPC, STS-B, and RTE tasks initialized with the parameters from the best MNLI model but do not apply any task-specific tricks to \textbf{WNLI} \citep[Winograd Schema Challenge NLI,][]{tasks_glue_wnli} and always predict the majority label.

\subsection{Results}
Table~\ref{tab:gluetest} presents the results on the test set  obtained from the GLUE evaluation server. Besides our best WARP models, we also include the human baselines, current state-of-the-art model \cite{deberta}, the regular fine-tuned pretrained model we use, and also include relatively small language models, including \cite{Jiao2020TinyBERTDB}, \cite{clark-etal-2020-pre}, \cite{Houlsby2019ParameterEfficientTL}.

\begin{table*}
\centering 
\small
\onehalfspacing
\setlength{\tabcolsep}{7pt}
\begin{tabular}{@{}lcccccccc|cr@{}}
\hline

& \textbf{MNLI}
& \textbf{QNLI}
& \textbf{QQP}
& \textbf{RTE}
& \textbf{SST}
& \textbf{MRPC}
& \textbf{CoLA}
& \textbf{STS-B}
& \multirow{2}{*}{\textbf{AVG}}
& \multirow{2}{*}{\textbf{\#} }
\\

\footnotesize{\textbf{train size}}
& \footnotesize{$392702$} & \footnotesize{$104743$} & \footnotesize{$363846$} & \footnotesize{$2490$} & \footnotesize{$67349$} & \footnotesize{$3668$} & \footnotesize{$8551$} & \footnotesize{$5749$} 
\\
\hline



Fine-Tuning
& $90.2$ & $94.7$ & $92.2$ & $86.6$ & $96.4$ & $90.9$ & $68.0$ & $92.4$ 
&  $88.9$ &$355 \cdot 10^6$ \\
Adapters
& $90.4$ & $94.7$ & $88.5$ & $83.4$ & $96.3$ & $92.9$ & $67.4$ & $92.5$ 
&  $88.3$ &$3 \cdot 10^6$ \\
Linear Classifier
& $64.2$ & $78.1$ & $74.9$ & $59.2$ & $88.4$ & $82.5$ & $48.9$ & $71.8$ 
&  $71.0$ &$\leq3072$ \\

\hline
 
WARP\textsubscript{0}
& $70.9$ & $78.8$ & $77.1$ & $72.2$ & $89.8$ & $83.8$ & $32.8$ & $73.8$ 
&  $72.4$ &$\leq3072$ \\
WARP\textsubscript{1}
& $83.9$ & $87.6$ & $81.6$ & $72.6$ & $93.8$ & $84.7$ & $46.1$ & $80.4$
&  $78.8$ &$\leq4096$ \\
WARP\textsubscript{2}
& $85.4$ & $88.0$ & $81.5$ & $69.7$ & $94.3$ & $85.3$ & $54.4$ & $80.8$
&  $79.9$ &$\leq5120$ \\
WARP\textsubscript{4}
& $86.9$ & $92.4$ & $83.1$ & $68.2$ & $95.9$ & $85.0$ & $56.0$ & $75.5$
&  $80.4$ &$\leq7168$ \\
WARP\textsubscript{8}
& $87.6$ & $93.0$ & $83.8$ & $72.9$ & $95.4$ & $85.6$ & $57.4$ & $81.0$
&  $82.1$ &$<11K$ \\
\hline
WARP\textsubscript{init}
& $86.8$ & $90.4$ & $83.6$ & $80.1$ & $96.0$ & $86.0$ & $51.7$ & $86.9$
&  $82.7$ &$<11K$ \\
WARP\textsubscript{20}
& \underline{$88.2$}
         & \underline{$93.5$}
                  & \underline{$84.5$}
 						   & $75.8$ & \underline{$96.0$}
 						                     & $90.8$ & \underline{$60.6$}
 						                                       & $88.6$ 
& 84.8 &$<25K$ \\
WARP\textsubscript{MNLI}
&        &        &        & \underline{$86.3$} &   
            								 & \underline{$91.2$}
            								 		  &        & \underline{$91.0$} 
& 86.4 &$<25K$ \\

\hline

\end{tabular}
\caption{
Dev set results on GLUE tasks. 
The last column shows the number of \textit{trainable} parameters only.
WARP\textsubscript{$i$} corresponds to WARP training with prompt consisting of $i$ prompt tokens.
WARP\textsubscript{MNLI} corresponds to WARP training initialized with the best MNLI parameters. All the models are based on pretrained \texttt{roberta-large}, and for Adapters and WARP-based approaches require to store $355 \cdot 10^6$ frozen parameters shared across all the GLUE tasks.
We show the primary validation metric for each task, described at Subsection \ref{section_glue_tasks}. The AVG column shows the average of shown metrics and is not comparable to the Test server GLUE Score.
The number of parameters for WARP methods may vary because of a difference in the number of classes. Underlined numbers correspond to our GLUE submission.
}
\label{tab:gluedev}
\end{table*}

With the GLUE Score, WARP outperforms all the models that train less than 25 million parameters on the leaderboard. We explain the relatively strong WARP results on textual entailment tasks by the easier reformulation of such tasks. Likewise, we explain the relatively weak performance on CoLA by the difficulties of reformulating the task into a Cloze task.

To further analyze WARP, we conduct several experiments and focus on dev set results.
In order to directly compare WARP with existing methods, we report in Table~\ref{tab:gluedev} different methods that use RoBERTa, including fine-tuning, linear classifiers on top, AutoPrompt, and Adapters.\footnote{Unlike in Table~\ref{tab:gluedev}, Adapters in Table~\ref{tab:gluetest} are built on \texttt{bert-large-uncased} model.} For WARP experiments, we compare performance with different numbers of prompt tokens. 

The WARP\textsubscript{0} model does not introduce any prompt parameters. The only difference between WARP\textsubscript{0} and Linear Classifier is that for WARP\textsubscript{0}, \texttt{[MASK]} is added to the input of each sample, and we get sentence representations from the MLM head at the masked position. By contrast, in the case of the Linear Classifier, we use the average of non-special token embeddings as sentence representations. As we can see, pooling with MLM is significantly better.

Table \ref{tab:gluedev} shows that, as we decrease the number of trainable prompt parameters, the performance decreases, but the model still works. Similar behavior was observed by \citet{Elsayed2018Adversarial} in experiments with different padding parameter sizes. However, in contrast to WARP, the number of trainable parameters in that work are much greater than the size of the input.

An important benefit of using WARP is that it can be initialized with manual prompts. In addition to the regular models where we initialize with \texttt{[MASK]} tokens, we performed a run on the GLUE datasets with the same prompt \pattern{\textsf{\small[CLS] ``}$S_1$\textsf{\small''?$\,\,$[MASK]. ``}$S_2$\textsf{\small''! [SEP]}} for all the tasks (without $S_2$ for single-sentence tasks). We denote these results as WARP\textsubscript{init} in Table \ref{tab:gluedev}. WARP\textsubscript{init} outperforms WARP$_8$ on tasks with relatively few training examples --- RTE, MRPC and STS-B, which indicates its potential in the low-data regime.

\section{Few-Shot Experiments}
\label{section_few_shot}
The fact that WARP can be initialized using manually designed natural prompts suggests that we can similarly benefit from such human attribution similar to iPET \cite{Schick2020ItsNJ}, especially in scenarios with limited training data.

\subsection{Setup}
For our few-shot experiments we build WARP on top of ALBERT \cite{albert}, the same pretrained model used by PET and iPET. To initialize WARP prompts, we use the same Prompt-Verbalizer Patterns (PVP) from \citet{Schick2020ItsNJ}: the embeddings for \texttt{[P\_1], [P\_2]... [P\_N]} are initialized with PVP's prompt token embeddings, and embeddings for \texttt{[V\_1], [V\_2]... [V\_C]} are initialized with verbalizer token embeddings for their corresponding classes.
Unlike \texttt{roberta-large}, the \texttt{alberta-xxlarge-v2} uses word embeddings of size 128 (8 times smaller than RoBERTa). 

\subsection{Tasks}
\label{section_few_shot_tasks}
In order to compare with GPT-3, PET, and iPET, we use two tasks from \textbf{FewGLUE} \cite{Schick2020ItsNJ}, which is a few-shot subset of the SuperGLUE benchmark \cite{tasks_super_glue} consisting of 32 examples for each task. The dataset also provides 20000 additional unlabeled examples, however, we do not make use of them and work in a purely supervised setup.

\textbf{CB}: CommitmentBank \cite{tasks_super_glue_cb} is a textual entailment task which we treat like the other sentence pair classification tasks. To initialize the prompt we use the template
\pattern{\textsf{\small[CLS] ``}$h$\textsf{\small''?$\,\,$[MASK]. ``}$p$\textsf{\small'' [SEP]}}.
We also initialize \texttt{[V\_1]}, \texttt{[V\_2]}, \texttt{[V\_3]} token embeddings with \texttt{\_yes}, \texttt{\_no} and \texttt{\_maybe}  (respectively for \texttt{entailment}, \texttt{contradiction} and \texttt{neutral}).

\textbf{RTE}: Unlike experiments on the RTE task for the full-sized training in the GLUE benchmark, we do not initialize the model with vectors from MNLI. Instead, the prompt is initialized exactly the same way as in the CB task. The only difference is that we have only the two tokens \texttt{[V\_1]} and \texttt{[V\_2]} initialized with  \texttt{\_yes} and \texttt{\_instead} (for \texttt{entailment} and \texttt{not\_entailment}, respectively).

\subsection{Model Selection}
Although all trainable parameters are manually initialized in this setup, different random seeds can yield different results because of the order the training examples appear during an epoch.

In the few-shot setup we cannot access  the original validation set. Thus, we disable early stopping and simply pick the last checkpoint.

In order to find the best initial learning rate, we conduct 20 runs of WARP with the same learning rate each time by randomly choosing 16 training examples and taking the rest for a development set. We repeat this for all candidate learning rates and choose the one with the best average validation performance across all the random seeds.

Finally, in order to eliminate the effect of different random seeds, we build an ensemble model from 20 WARP runs using simple majority vote.

\begin{table}
\centering
\begin{tabular}{clccc}
\hline
& \multirow{2}{*}{\textbf{Model}}
& \textbf{CB}
& \textbf{RTE}
\\
\small
& 
& F\textsubscript{1} / Acc.
& Acc.
\\
\hline

\multirow{6}{*}{\rotatebox{90}{dev}} 
& GPT-3 Small                        & $26.1$ / $42.9$ & $52.3$ \\
& GPT-3 Med                          & $40.4$ / $58.9$ & $48.4$ \\
& GPT-3                              & $57.2$ / $82.1$ & $72.9$ \\
& PET (ALBERT)                       & $59.4$ / $85.1$ & $69.8$ \\
& iPET (ALBERT)                      & $92.4$ / $92.9$ & $74.0$ \\
& WARP\textsubscript{init} (ALBERT)  & $84.0$ / $87.5$ & $71.8$ \\

\hline

\multirow{4}{*}{\rotatebox{90}{test}} 
& GPT-3                              & $52.0$ / $75.6$ & $69.0$ \\
& PET (ALBERT)                       & $60.2$ / $87.2$ & $67.2$ \\
& iPET (ALBERT)                      & $79.9$ / $88.8$ & $70.8$ \\
& WARP\textsubscript{init} (ALBERT)  & $70.2$ / $82.4$ & $69.1$ \\

\hline

\end{tabular}

\caption{Results on SuperGLUE benchmark. The results for the test set are obtained from SuperGLUE evaluation server.
We only show systems performing in a similar few-shot training setup using 32 examples. 
}
\label{tab:supergluetest}

\end{table}

\subsection{Results}
As seen in Table \ref{tab:supergluetest}, WARP outperforms PET and GPT-3 baselines, but stays behind iPET on both tasks. GPT-3 has 170B parameters, but none of them is being trained for the given tasks. PET and iPET have 255M parameters, and \textit{all} of them are trained for these tasks. Additionally, they leverage unlabeled examples using distillation. WARP has roughly the same 255M parameters, but only 1024 of them are trained for any single model. An ensemble of 20 WARP models has slightly more than 20K trainable parameters.


\section{Discussion}\label{sec:discussion}

\subsection{Interpreting tokens learned by WARP}
WARP learns prompt embeddings in a continuous space. In this section, we explore those embeddings by looking at the nearby token vectors. Table \ref{tab:tokens} in the Supplementary material lists the closest tokens (in terms of cosine similarity) to the learned embeddings. All GLUE tasks are initialized with \texttt{[MASK]} token, except for RTE, MRPC, and STS-B, which are initialized from the pretrained MNLI model. The prompt tokens of the solutions for those three tasks are quite close to the ones from the MNLI solution. We have seen similar behavior on SuperGLUE experiments with manual initializations. The solution for CoLA (which is one of the worst-performing tasks) is close to the initialized point. 

We do not see any prompt tokens that are meaningful in the context of the tasks. As expected, the verbalized tokens are more interpretable. For example, the embedding for the ``contradiction'' class of MNLI is close to the token ``Unless''. The embeddings for ``negative'' and ``positive'' classes of SST-2 task are close to ``defective'' and ``important'', respectively. Other verbalized tokens are non-interpretable (e.g. ``470'' or word pieces with non-Latin characters). 


\subsection{Comparison with AutoPrompt}
\label{sec:autoprompt}

AutoPrompt \cite{shin-etal-2020-autoprompt} learns a prompt for the given task in the finite space of vocabulary tokens. Their best version uses 3 or 6 prompt tokens and reaches 91.2\% accuracy on the development set of SST-2. The search space of WARP is significantly larger, which allows WARP to get better performance with just a single prompt token (93.8\%).

AutoPrompt does not achieve meaningful results on RTE or CB tasks. WARP succeeds on both without manual initialization. Moreover, with manual initialization, WARP gets good performance on both tasks even with just 32 examples (Table \ref{tab:supergluetest}).

\begin{figure}
    \centering
    \includegraphics[width=\columnwidth]{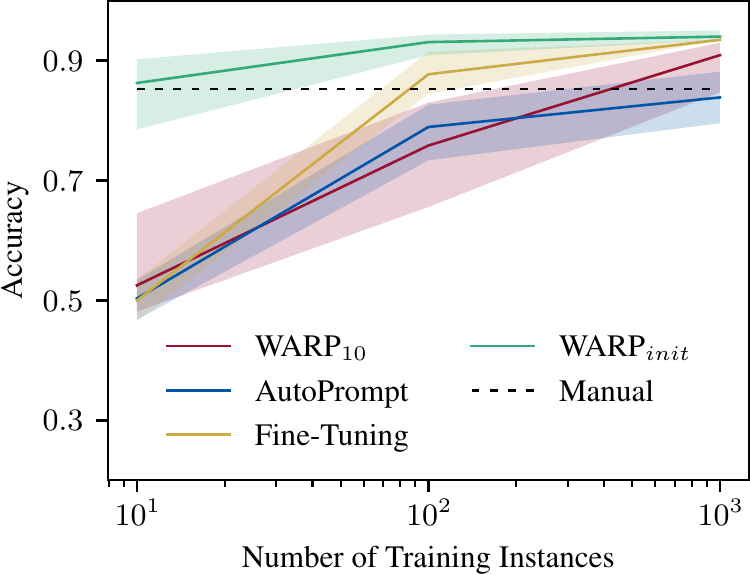}
    \caption{The effect of the training data size for SST-2 task (dev set). Horizontal axis is the number of training examples. Solid lines represent median over 10 runs, and the error bars show minimum and maximum performance. All methods use \texttt{roberta-large} model. The results for AutoPrompt and fine-tuning are taken from \cite{shin-etal-2020-autoprompt}}
    \label{fig:data_size}. 
\end{figure}

Figure \ref{fig:data_size} shows the dependence of the accuracy on SST-2 development set from the number of training samples. Both WARP and AutoPrompt use 10 prompt tokens. With a few hundred training samples or fewer, the difference between the two algorithms is not significant. WARP starts to perform better with more training samples.

\citet{shin-etal-2020-autoprompt}  include results  with a manually designed prompt\footnote{
\pattern{\textsf{\small \emph{SENT.} this movie was \mask{}.}} as a prompt, and ``terrible'' and ``fantastic'' as verbalizer tokens} which performs pretty well (shown as a dashed line). We also compare with the manually initialized\footnote{\pattern{\textsf{\small \emph{SENT}, and finally, the movie overall was very \mask{}! }} as a prompt, and ``good'' and ``bad'' as verbalizer tokens} version of WARP, which performs very well with just 100 examples.

\subsection{Real-world applications}
\label{section_real_world}
The importance of NLP systems like WARP can be demonstrated by the following application. Suppose we want to build a system that needs to serve $N >> 1$ classification tasks simultaneously. Let the number of classes for each task be bounded by $C$. The system can be based on a large pretrained language model with $M$ parameters, using word embedding size $E$. How many parameters should the system store in the device memory to be able to serve all $N$ tasks? 

If we take the approach with frozen features, we can reuse $M$ parameters for all tasks and store additional $ECN$ task-specific parameters. This is optimal in terms of storage but will not perform well. The other extreme is to fine-tune the whole model for each task and store at least $MN$ parameters. Table \ref{tab:params} shows the trade-offs offered by the other solutions. Methods like TinyBERT \textit{decrease} the number of parameters from $MN$ by only $M$. WARP, on the other hand, needs to \textit{store} only $M + NE(C+K)$ parameters, where $K$ is the number of trainable prompt tokens. 

\begin{table}[]
\begin{tabular}{ll}
\hline
\textbf{Approach}                 & \textbf{\# of parameters to store}  \\\hline
Linear probing   & $M+ECN$       \\
Full fine-tuning & $MN$          \\
Single layer     & $M+NE(E+C)$   \\
TinyBERT         & $M_0N$        \\
Adapters         & $M + NEE'$    \\
WARP             & $M + NE(C+K)$ \\ \hline
\end{tabular}
\caption{The number of parameters to be stored to serve $N$ text classification tasks with at most $C$ classes each, using a pretrained language model with $M$ parameters. $E$ is the dimension of embeddings ($1024$ in the case of RoBERTa). In TinyBERT, $M_0$ can be up to 10 times less than $M$. In Adapters, $E'$ is roughly equal to $E$, as the number of layers to which adapters are attached roughly compensates the smaller size of the bottleneck layer. In WARP, $K$ is the number of prompts  (usually fewer than 10).}\label{tab:params}
\end{table}

In practice, WARP additionally allows performing inference on inputs for different tasks in parallel, using samples of multiple tasks in the same batch.
Every input sentence can be concatenated with task-specific pretrained prompts in advance. Then, the forward pass of the network is identical for all tasks. The final task-specific linear layers can be concatenated to form a single large linear layer with at most $NC$ output neurons.

This approach can be especially useful in the systems that provide machine learning models as a service. By storing one copy of a pretrained language model, it is possible to serve a large number of user-specific models in parallel with little overhead. 

\section{Conclusion}
In this paper we have proposed an alternative way to transfer knowledge from large pretrained language models to downstream tasks by appending carefully optimized embeddings to the input text. The method outperforms existing methods with significantly more trainable parameters on GLUE benchmark tasks and shows an impressive performance in a few-shot setting on two SuperGLUE tasks. On the sentiment analysis task, the performance is comparable to the fully fine-tuned language models. This method can save a lot of storage in software applications designed to serve large numbers of sentence classification tasks. 

\section*{Acknowledgments}

This work is based in part on research sponsored by Air Force Research Laboratory (AFRL) under agreement number FA8750-19-1-1000. The U.S. Government is authorized to reproduce and distribute reprints for Government purposes notwithstanding any copyright notation therein. The views and conclusions contained herein are those of the authors and should not be interpreted as necessarily representing the official policies or endorsements, either expressed or implied, of Air Force Laboratory, DARPA or the U.S. Government.

The work was supported by the RA Science Committee, in the frames of the research project No. 20TTAT-AIa024. Most experiments were performed on GPUs donated by NVIDIA. 
\bibliography{anthology,acl2020}
\bibliographystyle{acl_natbib}

\clearpage

\appendix

\section{Hyperparameters}
For each of the tasks, we performed hyperparameter search in the following space:
\begin{itemize}
  \item \textbf{Learning rate} is chosen from the set $\{10^{-2}, 3\cdot10^{-3}, 10^{-3}, 3 \cdot 10^{-4}, 10^{-4}, 3 \cdot 10^{-5}\}$,
  \item \textbf{Number of epochs} is chosen as either $10$ or $20$. This determines the behavior of the slanted triangular learning rate scheduler.
  \item \textbf{Initialization} is performed either with the embedding of the \texttt{[MASK]} token, or randomly initialized from a normal distribution, with the mean and variance taken from the matrix of RoBERTa's word embeddings.
\end{itemize}

The hyperparameter search took roughly 4 days on two Titan V GPUs. The final choices for each task are shown in Table \ref{hparams}. Initialization with \texttt{[MASK]} performed better than the random initialization.

We disable all dropouts inside Transformer. We use \textit{huggingface} implementation of AdamW optimizer with weight decay disabled. The gradient is normalized to the value $1.0$.
For the batch sampling we use bucketing with padding noise of $0.1$. In order to use the device memory more effectively, we also set maximum number of tokens per batch to $2048$. The maximum sequence length is truncated to $512$ tokens.
We enable mixed precision and pad all sequence lengths to the multiples of $8$ for the effective usage of TensorCores\footnote{https://docs.nvidia.com/deeplearning/performance/mixed-precision-training/index.html}.

\begin{table}[b]
\begin{tabular}{cccc}
\hline
    \textbf{Task}  & \textbf{Learning rate} & \textbf{Epochs} & \textbf{Init.}   \\
\hline
MNLI      & $0.001$    & $10$        & [MASK] \\
QNLI      & $0.001$    & $10$        & [MASK] \\
QQP       & $0.0003$   & $20$        & [MASK] \\
RTE       & $0.001$    & $20$        & MNLI   \\
SST-2     & $0.003$    & $20$        & [MASK] \\
MRPC      & $0.001$    & $20$        & MNLI   \\
CoLA      & $0.001$    & $20$        & [MASK] \\
STS-B     & $0.001$    & $20$        & MNLI   \\
\hline
\end{tabular}

\caption{Hyperparameters of our best-performing models. \texttt{[MASK]} means the prompts are intialized with the word embedding of same token, and MNLI means the prompt is initialized with the prompts of out best MNLI run.}
\label{hparams}
\end{table}

\section{Learned Tokens}
Table \ref{tab:tokens} lists the closest vocabulary words to the learned embeddings. Most tasks have two input sentences, so the prompts consist of three parts: one is added before the first sentence, the second one is added between the sentences and the third one is appended next to the second sentence. For the single-sentence tasks, the second and third parts of the prompt are simply concatenated. Each task has trainable verbalizer tokens, one per output class.

The prompts of RTE, MRPC and STS-B are pretty similar to MNLI's prompts, as the models for these tasks were initialized from pretrained MNLI models. The other tasks were initialized with \texttt{[MASK]} tokens. The final model for CoLA didn't move too far from its initialization.


\begin{table*}[]
\centering
\begin{tabular}{|c|c|c|p{9cm}|}

\hline

\multirow{6}{*}{\textbf{MNLI}}
            &  \multirow{3}{*}{\textbf{Prompts}}        & \textbf{before}             & \small A-A-A-A-A-A-A-A-A-A-A-A-A-A-A-A-        Tomorrow                                Ale                                     .aGj                                    \_*.                                   \\
\cline{3-4} &                                            & \textbf{between}            & \small \_MUCH                                   irin                                    {[}/                                    \_a                                      \_(@                                     \textbf{[MASK]}   \_dL                                     aHJ                                     E                                       [MASK]                                  \_aKH                                  \\
\cline{3-4} &                                            & \textbf{after}              & \small \_\textless{}!--                         \_informing                              inyl                                    \_entit                                  dim                                   \\

\cline{2-4} & \multirow{3}{*}{\textbf{Verbalizers}}      & \textbf{entailment}         & \small \_categories    \\
\cline{3-4} &                                            & \textbf{neutral}            & \small gomery    \\
\cline{3-4} &                                            & \textbf{contradiction}      & \small Unless    \\
\hline
\hline

\multirow{5}{*}{\textbf{QNLI}}
            &  \multirow{3}{*}{\textbf{Prompts}}        & \textbf{before}             & \small *.                                      \_neigh                                  [MASK]                                  U                                       \_\{\{                                 \\
\cline{3-4} &                                            & \textbf{between}            & \small aG|aG|                                  [MASK]                                  olitan                                  \_pronouns                               [MASK]                                  \textbf{[MASK]}   [MASK]                                  @@@@                                    [MASK]                                  \_Choi                                   [MASK]                                \\
\cline{3-4} &                                            & \textbf{after}              & \small \\

\cline{2-4} & \multirow{2}{*}{\textbf{Verbalizers}}      & \textbf{entailment}         & \small \_VIDE    \\
\cline{3-4} &                                            & \textbf{not\_entailment}     & \small 470    \\
\hline
\hline

\multirow{5}{*}{\textbf{QQP}}
            &  \multirow{3}{*}{\textbf{Prompts}}        & \textbf{before}             & \small \_resembling                             \_swarm                                  \_Paramount                              \_Calm                                   \_Membership                           \\
\cline{3-4} &                                            & \textbf{between}            & \small \_derive                                 rics                                    [MASK]                                  \_alias                                  iary                                    \textbf{[MASK]}   \_omnip                                  [MASK]                                  [MASK]                                  [MASK]                                  \_sham                                 \\
\cline{3-4} &                                            & \textbf{after}              & \small [MASK]                                  \_forb                                   [MASK]                                  \_Firefly                                \_THEY                                 \\

\cline{2-4} & \multirow{2}{*}{\textbf{Verbalizers}}      & \textbf{not\_duplicate}         & \small ende    \\
\cline{3-4} &                                            & \textbf{duplicate}            & \small \_sugg    \\
\hline
\hline

\multirow{5}{*}{\textbf{RTE}}
            &  \multirow{3}{*}{\textbf{Prompts}}        & \textbf{before}             & \small A-A-A-A-A-A-A-A-A-A-A-A-A-A-A-A-        Tomorrow                                ALE                                     .aGj                                    \_*.                                   \\
\cline{3-4} &                                            & \textbf{between}            & \small \_MUCH                                   irin                                    {[}/                                    \_a                                      \_(@                                     \textbf{[MASK]}   \_                                      aHJ                                     femin                                   [MASK]                                  \_aK                                   \\
\cline{3-4} &                                            & \textbf{after}              & \small ahiahi                                  \_informing                              \#                                   \_entit                                  OOOO                                  \\

\cline{2-4} & \multirow{2}{*}{\textbf{Verbalizers}}      & \textbf{entailment}         & \small e!    \\
\cline{3-4} &                                            & \textbf{not\_entailment}            & \small \_blames    \\
\hline
\hline

\multirow{5}{*}{\textbf{SST-2}}
            &  \multirow{3}{*}{\textbf{Prompts}}        & \textbf{before}             & \small choes                                   \_charms                                 \_sorely                                 \_"...                                   akijakij                                \\
\cline{3-4} &                                            & \textbf{between}            & \small a                                       afe                                     Pae                                      \_charred                                \_masked                                 \textbf{[MASK]}   \_Fall                                   \_babys                                  \_smartest                               ik                                      /                                     \\
\cline{3-4} &                                            & \textbf{after}              & \small dL                                      forums                                  \_bio                                    \_mang                                   A+-                                    \\

\cline{2-4} & \multirow{2}{*}{\textbf{Verbalizers}}      & \textbf{negative}           & \small \_defective    \\
\cline{3-4} &                                            & \textbf{positive}           & \small \_important    \\
\hline
\hline

\multirow{5}{*}{\textbf{MRPC}}
            &  \multirow{3}{*}{\textbf{Prompts}}        & \textbf{before}             & \small A-A-A-A-A-A-A-A-A-A-A-A-A-A-A-A-        Tomorrow                                rison                                   .aGj                                    \_*.                                   \\
\cline{3-4} &                                            & \textbf{between}            & \small  \_MUCH                                   irin                                    {[}/                                    \_a                                      jay                                     \textbf{[MASK]}   \_dL                                     aHJ                                     femin                                   [MASK]                                  .?                                    \\
\cline{3-4} &                                            & \textbf{after}              & \small  \_\textgreater{}                        \_informing                              \#                                   \_entit                                  OOOO                                  \\
\cline{2-4} & \multirow{2}{*}{\textbf{Verbalizers}}      & \textbf{entailment}         & \small \_categories    \\
\cline{3-4} &                                            & \textbf{neutral}            & \small gomery    \\
\hline
\hline

\multirow{5}{*}{\textbf{CoLA}}
            &  \multirow{3}{*}{\textbf{Prompts}}        & \textbf{before}             & \small [MASK]                                  [MASK]                                  [MASK]                                  [MASK]                                  [MASK]                                \\
\cline{3-4} &                                            & \textbf{between}            & \small [MASK]                                  [MASK]                                  [MASK]                                  [MASK]                                  [MASK]                                  \textbf{[MASK]}   [MASK]                                  [MASK]                                  [MASK]                                  [MASK]                                  [MASK]                                \\
\cline{3-4} &                                            & \textbf{after}              & \small [MASK]                                  [MASK]                                  [MASK]                                  [MASK]                                  [MASK]                                \\

\cline{2-4} & \multirow{2}{*}{\textbf{Verbalizers}}      & \textbf{unacceptable}         & \small \_additionally    \\
\cline{3-4} &                                            & \textbf{acceptable}            & \small o    \\
\hline
\hline

\multirow{4}{*}{\textbf{STS-B}}
            &  \multirow{3}{*}{\textbf{Prompts}}        & \textbf{before}             & \small A-A-A-A-A-A-A-A-A-A-A-A-A-A-A-A-          Tomorrow                                  Ale                                       .aGj                                      [MASK]   \\
\cline{3-4} &                                            & \textbf{between}            & \small \_Kers                                     irin                                      {[}/                                      \_a                                        \_(@                                       \textbf{[MASK]}     \_dL                                       AhAHAhAH                                  femin                                     [MASK]     \_aKH                                     \\
\cline{3-4} &                                            & \textbf{after}              & \small A-A-A-A-A-A-A-A-A-A-A-A-A-A-A-A-          \_repertoire                               inyl                                      \_Idea                                     dim                                   \\
\cline{2-4} & \multirow{1}{*}{\textbf{Verbalizers}}      & \textbf{regression}         & \small cH    \\
\hline
\hline

\end{tabular}
\caption{The closest words to the prompt and verbalizer token embeddings for the best model for each task. We use cosine distance to measure the distance. \textbf{[MASK]} tokens highlighted in bold indicate the positions we use to output the prediction. }
\label{tab:tokens}
\end{table*}


\end{document}